\documentclass[10pt,journal,compsoc]{IEEEtran}

\ifCLASSOPTIONcompsoc
  \usepackage[nocompress]{cite}
\else
  \usepackage{cite}
\fi
\ifCLASSINFOpdf
\else
\fi

\usepackage{framed,multirow}
\usepackage{graphicx}
\usepackage{textcomp}
\usepackage{hyperref}
\usepackage{amssymb}
\usepackage{latexsym}
\usepackage{algorithmic}
\usepackage{url}
\usepackage{xcolor}
\usepackage{cite}
\usepackage{ragged2e}

\begin{document}
%
\title{A Cross-Modal Image Fusion Method Guided by Human Visual Characteristics}
%
%
%
%

\author{\IEEEauthorblockN{Aiqing Fang,
		Xinbo Zhao\IEEEauthorrefmark{1},
		Jiaqi Yang,
		Yanning Zhang, \\
		jiaqi Yang,Yanning Zhang,~\IEEEmembership{Senior Member,~IEEE}}\\
	\IEEEauthorblockA{\justifying National Engineering Laboratory for Integrated Aero-Space-Ground-Ocean Big Data Application Technology,School of Computer Science and Engineering, Northwestern Polytechnical University, Xi’an 710072, China.}
	\thanks{Corresponding author: Xinbo Zhao (email: xbozhao@nwpu.edu.cn)
}}

\IEEEdisplaynontitleabstractindextext

%
%

\markboth{IEEE TRANSACTIONS ON MULTIMEDIA}%
{Shell \MakeLowercase{\textit{et al.}}: Bare Advanced Demo of IEEEtran.cls for IEEE Computer Society Journals}
%



\IEEEtitleabstractindextext{%
\begin{abstract}
	\justifying
The characteristics of feature selection, nonlinear combination and multi-task auxiliary learning mechanism of the human visual perception system play an important role in real-world scenarios, but the research of image fusion theory based on the characteristics of human visual perception is quite limited. Inspired by the characteristics of human visual perception, we propose a multi-task auxiliary learning optimization image fusion method. \textit{For the first time, we explore the impact of human visual information characteristics on image fusion tasks in the field of image fusion. Besides, our method provides a new solution to the problem that cross-modal image fusion loss function is difficult to model.} \textbf{Firstly}, we combine channel attention model with nonlinear convolutional neural network to select effective features and fuse nonlinear weight. \textbf{Secondly}, we analyze the impact of the existing image fusion loss on the image fusion quality, and establish the multi-task loss function model of semi-supervised learning network. \textbf{Thirdly}, aiming at the multi-task auxiliary learning mechanism of human visual perception system, we study the influence of multi-task auxiliary learning mechanism on image fusion task based on the single task with multi-loss network. By simulating the three characteristics of human visual perception system, the fused image is more consistent with the mechanism of human brain image fusion. \textbf{Finally}, in order to verify the superiority of our algorithm, we carried out experiments on the combined vision system image data set, and extended our algorithm to the infrared and visible image and the multi-focus image public data set for experimental verification. The experimental results demonstrate the superiority of our image fusion method.
\end{abstract}

\begin{IEEEkeywords}
Image fusion, auxiliary learning, nonlinear fusion characteristics, deep learning.
\end{IEEEkeywords}}

\maketitle

\IEEEdisplaynontitleabstractindextext

%
\IEEEpeerreviewmaketitle

\ifCLASSOPTIONcompsoc
\IEEEraisesectionheading{\section{Introduction}\label{sec:introduction}}
\else
\section{Introduction}
\label{sec:introduction}
\fi

The human visual perception system has better performance than the existing derived algorithms in object detection \cite{HongliangLi2013CODF}\cite{LiuDi2020ACFN}, image caption \cite{DongJianfeng2018PVFF}, object tracking and other tasks. Therefore, \textit{we believe that the human vision system is also robust to image fusion tasks}. \textit{According to the research of cognitive psychology and neuroscience theory \cite{KochC1985Sisv,Treisman1980A,MillerCPortex}, the information processing of human visual perception system has the characteristics of feature selection, nonlinear combination and multi-task auxiliary learning mechanism.} We believe that \textit{based on these three characteristics of human brain, human brain is more robust than existing derivative algorithms in image fusion task or other visual task processing tasks, as verified in Sect.4.} 

However, in the task of image fusion, the existing image fusionmethods have paid few research attention on the characteristics of human visual perception system. By contrast, more researches have been conducted from the technology perspective, without considering the characteristics of human visual perception and information processing mechanism. We can divide image fusion methods into traditional image fusion methods and image fusion methods based on deep learning. Traditional image fusion methods mainly include multi-scale transformation \cite{Zhao2017Multisensor,Cui2015Detail,ZhangInfrared}\cite{Bhatnagar} and visual significance \cite{Bavirisetti2016Two,Zhang2015A,Liu2017InfraredJSR-SD,Ma2017InfraredWLS,Cui2015Detail,ZhangInfrared} et al. \textit{There are three main approaches to image fusion based on deep learning.} \textit{1) The first one} is the combination of image transformation and deep learning feature, which only uses the convolutional neural network model of pretraining to provide deep learning feature maps \cite{Lahoud2019FastZERO,Li_2018DL,Liu2017InfraredCNN,Zhang_2018RCAN,Ma2018Infrared}\label{method}. \textit{2) The second} is the end-to-end convolution neural network method based on the twin networks, which uses the objective function for iterative optimization learning strategy \cite{Li2018DenseFuse,PrabhakarK.Ram2017DADU,YanXiang2018UDMI,MaFusionGAN}. \textit{3) The third} is to build an end-to-end deep convolution neural network (CNN), which is different from the second one in that it transforms image fusion into image classification. This method is more applicable to multi-focus image fusion tasks \cite{Liu2017MultiCNN,MaBoyuan2019SAUD}. \textit{In terms of image fusion criteria}, both traditional image fusion algorithm and deep learning image fusion method mainly adopt maximum fusion, sum fusion and weighted average fusion \cite{Ma2018Infrared,Lahoud2019FastZERO,Li_2018DL,Liu2017InfraredCNN,Zhang_2018RCAN,PrabhakarK.Ram2017DADU}. From the perspective of the universality of image fusion, a general image fusion framework (IFCNN)
proposed by Zhang\cite{ZhangYu2020IAgi} based on multi-exposure fusion (MEF)
framework \cite{PrabhakarK.Ram2017DADU}. This method uses supervised learning method in multi-focus data set, and then applies training weight to different image fusion tasks according to different fusion rules. This method focuses more on the generality of fusion framework than on the robustness of fusion algorithm. 

\begin{figure*}[ht]
	\centering
	\includegraphics[scale=3.5,width=0.9\textwidth]{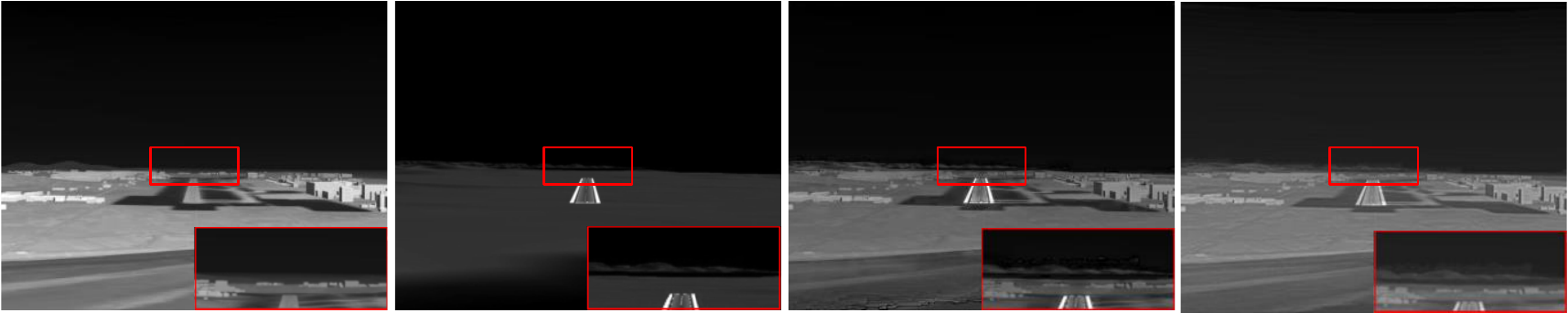}
	\caption{Combined vision system (CVS) image fusion. From left to right: the enhanced vision system image (EVS), the synthetic vision system image (SVS), the fusion result of a classic method, and the fusion result of ours. Our method has a good fusion effect for image details, and the fusion effect is more coincident with the human visual perception mechanism.\label{summary}}
\end{figure*}

Although the existing image fusion theory has made remarkable achievements, it still has a big gap with the human brain image fusion effect. In order to make the results of image fusion more consistent with the mechanism of human brain image fusion and narrow the gap between human brain image fusion and human brain image fusion, we propose an cross-modal image fusion method based on human visual perception characteristics. In order to demonstrate the superiority of our fusion method in existing mainstream algorithm, we give a representative example in Fig. \ref{summary}. We use the EVS and SVS images in CVS image fusion data set to do qualitative comparison experiments. The two images on the left are EVS image and SVS image, the third image is the fusion effect of traditional operator, and the last image is the fusion result of our algorithm. From the fusion images of different algorithms, we can see that our algorithm is superior to the mainstream image fusion algorithm in subjective vision.

\textit{Our image fusion method uses multi-tasks loss function to optimize image fusion weight, and through the effective combination of attention mechanism and nonlinear neural network to simulate the feature selection characteristics and nonlinear combination characteristics of human brain image fusion}, which effectively improves the robustness of image fusion. The image fusion method we proposed is not a simple combination of existing deep learning methods. In the task of cross-modal image fusion, we make full use of the strong feature representation ability and nonlinear fitting ability of convolutional neural network, We use network architecture to simulate the characteristics of human visual system.

The \textit{main contributions} of our work include the following four points:

\textbf{Firstly}, we analyze the nonlinear characteristics and feature selection characteristics of human vision system, simulate the characteristics, and introduce the characteristics into the image fusion task.

\textbf{Secondly}, we introduce human brain auxiliary learning mechanism into image reconstruction task and multi-focus image fusion task, and study the influence of auxiliary learning mechanism on image fusion task.

\textbf{Thirdly}, based on the above theoretical research, we propose a robust multi-task loss function collaborative optimization learning image fusion method. To a certain extent, the method overcomes the difficulty of modeling the objective function of the cross-modal image, and provides a new fusion idea for the cross-modal image fusion task through the cooperative optimization of multi-task loss.

\textbf{Finally}, code and dataset. In order to speed up the research progress of researchers in the field of image fusion, more than 20 latest image fusion algorithm codes and 8 image fusion algorithm codes that have not been compared in this paper will be summarized on my \href{https://github.com/AiqingFang/Image-Fusion-Summary}{GitHub} homepage after paper is accepted.

The rest of this paper is laid out as follows. In section 2, we discusses three characteristics of human visual perception system. In section 3, we will analyze our proposed image fusion method based on the three characteristics of human visual perception system. In section 4, experiments and results analysis of different algorithms on different public datasets. And the results are qualitatively analyzed and discussed in section 5, and the experimental conclusions are summarized in section 6.


\section{RELATED WORK}
In this section, we will review the development of image fusion and inspired chacteristics.

\subsection{Image fusion}
Here we classify the cross-modal image fusion algorithms into traditional image fusion algorithm and deep learning method. We will review the representative algorithms in these fusion algorithms. 

\textit{1) Image fusion method based on traditional method}. Durga et al. \cite{Bavirisetti2016Two} proposed an image fusion method combining two-scale transformation and bottom-up visual saliency model. Zhang et al. \cite{ZhangInfrared} evaluated local edge-preserving (LEP) filter and saliency analysis to retain the details of a visible image with a discernible object area. Zhao et al. \cite{Zhao2017Multisensor} evaluated the effectiveness of decomposition components can be modeled by tailed $\alpha$ stable-based random variable distribution when multisensor image fusion. Kou et al. \cite{KouFei2018IDEf} introduced an edge-preserving smoothing pyramid in multi-scale exposure image fusion task. Lahoud et al. \cite{Lahoud2019FastZERO} proposed the image fusion method of meta learning, which combines multi-scale transformation with deep learning features and saliency features map of pretraining model. The above methods analyze and study image fusion from multi-scale decomposition and visual saliency weight. 

\textit{2) Image fusion method based on deep learning}. Liu et al. \cite{Yu2017medicalCNN} introduced deep learning method into the task of cross-modal image fusion task, which proves the effectiveness of deep learning method. PrabhakarK et al. \cite{PrabhakarK.Ram2017DADU} proposed an end-to-end deep learning image fusion. Bin et al. \cite{XiaoBin2020MIFb} evaluated the effective of hessian matrix in multi-focus image fusion task. At the same time, Ma et al. \cite{MaFusionGAN} introduced the generative adversarial network into infrared and visible image fusion for the first time. However, the image fusion effect of this method is fuzzy and smooth, lacking of rich texture information. To overcome this problem, Ma et al. \cite{MA202085} presented a detail preserving learning framework. The detail information of fusion image is effectively preserved by detail loss and edge loss. Zhang et al. \cite{ZhangYu2020IAgi} proposed a general image fusion task and the influence of different fusion criteria on image fusion performance is explored. 

In the task of cross-modal image fusion, the image fusion algorithm based on deep learning is better than the traditional image fusion algorithm and meta learning method. However, the unsupervised learning method without ground truth labels still has a big gap in image fusion quality compared with traditional method and meta learning method. There are two main reasons. \textbf{Firstly}, because the traditional image fusion algorithm or meta learning method is an artificial design of image fusion process, \textit{to a certain extent, it adds the subjective prior knowledge of the image}, so this method can achieve better fusion effect than unsupervised deep learning method without ground truth; \textbf{Secondly}, the fusion effect of deep learning image fusion method is closely related to the definition of loss function.


\subsection{Inspired chacteristics}
Although deep learning method has a very strong advantage in feature extraction, there is no prior knowledge such as ground truth labels or appropriate loss function, the effect of image fusion will be very poor. However, the prior knowledge is closely related to the human visual system. According to the research of Koch \cite{KochC1985Sisv} on visual selective attention, the classical theory of feature integration in cognitive psychology \cite{Treisman1980A} and \cite{MillerCPortex} , \textit{the processing of information in human visual system has the characteristics of feature selection, nonlinear combination. Human brain has the characteristics of multi-task auxiliary learning mechanism.} As a highly complex nonlinear system, human brain system will filter the perceptual target features based on subjective intention, ignore uncertain signals, and fuse the non mutually exclusive features that meet the subjective needs according to prior knowledge \cite{Treisman1980A}. 

\textit{1) In the aspect of feature selection and nonlinear fusion}, Hu et al. \cite{hu2017squeezeandexcitation} proposed senet channel attention network for image classification task in order to explicitly model the interdependence between different feature channels. The core idea of this method is to guide the network to learn the effectiveness of different channel features through loss function, give more weight to effective feature channels, and give less weight to unimportant feature channels. In the image fusion task, Fang et al. \cite{fang2019robust}\cite{fang2019crossmodal} first explored the influence of feature selection and nonlinear characteristics of human vision system on image fusion task.

\textit{2) In the aspect of multi-task auxiliary learning}, the human brain will reason and explore the new target task according to the prior knowledge stored in the brain memory, so as to complete the perception of the new task. In recent years, although deep learning method has made a breakthrough in many fields, there is a big gap between this method and human perception system. The existing network is more about the learning perception of the fixed situation task, unable to learn more and more complex tasks quickly, and human beings can complete them quickly. In view of the mechanism of multi-task auxiliary learning, Yang et al. \cite{GuangYang2009Smds} first proposed to use the important parameters in the prior knowledge based on the synaptic integration method, and modify the unimportant parameters according to the task \cite{KirkpatrickJames2017Ocfi}. For the problem of "catastrophic forgetting" in the process of multi-task auxiliary learning and the problem of increasing storage space in existing methods, Zeng et al. \cite{ZengGuanxiong2018CLoC} presented an algorithm of orthogonal weight modification and dynamic situation awareness processing. In view of the problem of weight balance in multi-task auxiliary learning, Kendall et al. \cite{KendallAlex2018MLUU} proposed the weight of different task losses based on the same variance uncertainty in Bayesian model. This method has achieved good results in three tasks, scene segmentation, instance segmentation and depth estimation. However, the relevant work has not been found in the field of image fusion.

Inspired by the three characteristics of human visual system and the existing research theories of the three characteristics, we propose the image fusion method guided by human visual characteristics. \textit{The method combines channel attention module and nonlinear convolution neural network to simulate the feature selection and nonlinear combination of human visual characteristics.} At the same time, we introduce the mechanism of multi-task auxiliary learning into the existing image fusion method, and \textit{use multi-task network to assist the main task of image fusion.}

\begin{figure*}[ht]
	\centering
	\includegraphics[width=0.7\textwidth]{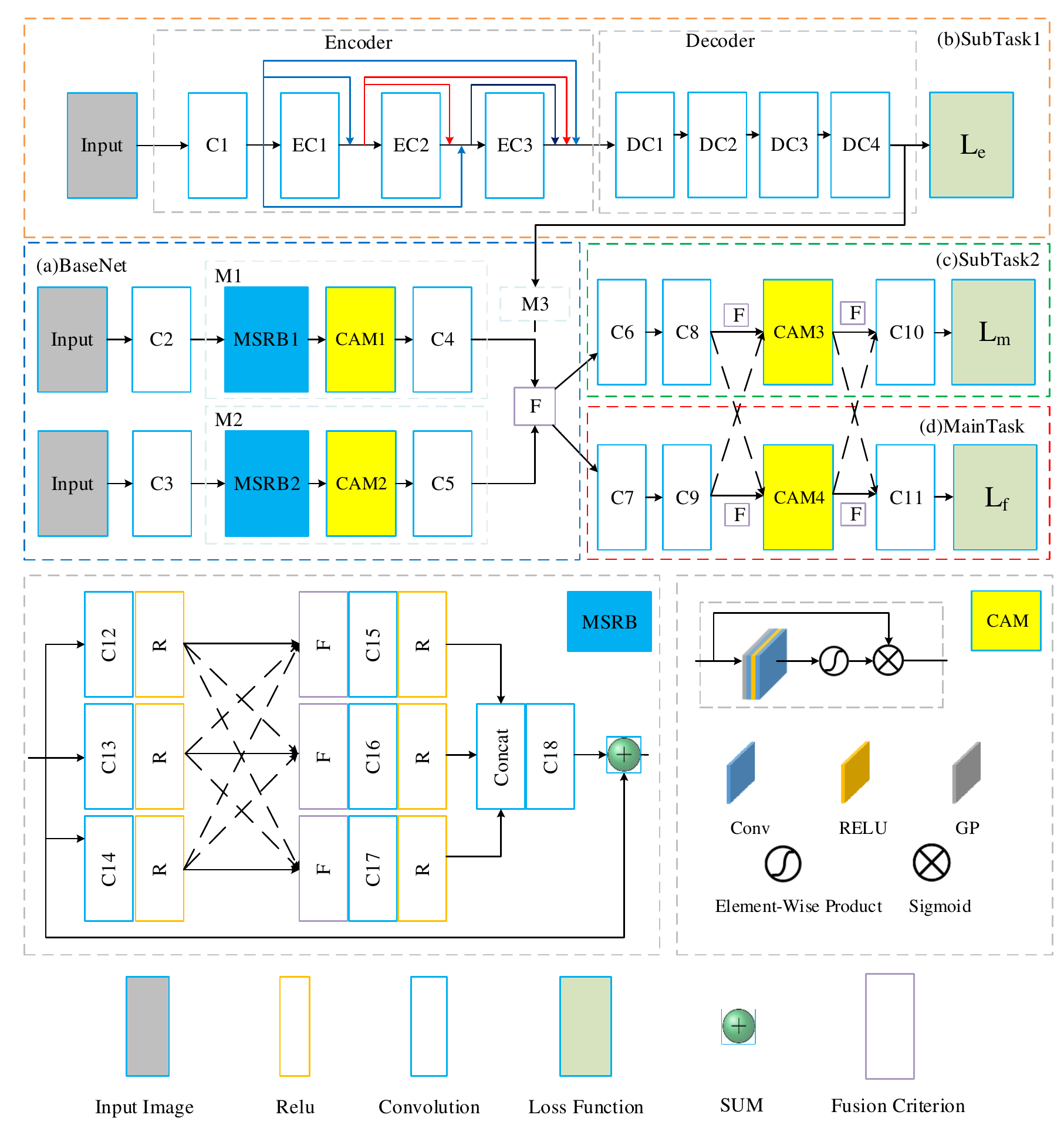}
	\caption{Unsupervised network architecture of ALAN. Where $L_m$ represents the loss of multi-focus image fusion task; $L_f$ represents the loss of CVS fusion task; $L_e$ represents the loss of image reconstruction task; $M$ represents the combination of MSRB, CAM and convolution. The orange dash box is subtask 1 module. Blue dashed frame is the basic backbone network of image fusion main task module and subtask 2 module. The green dash box is the network of subtask 2. Red dash is the main task network module of image fusion.\label{general}}
\end{figure*}

\section{METHOD}
As shown in Fig. \ref{general}, it is an semi-supervised multi-task auxiliary learning image fusion framework ALAN based on our proposed image fusion method. Our network framework mainly consists of a main network and two subnetworks. The main task network is mainly used for cross-modal image fusion task, and the two subnetworks are image reconstruction task network and multi-focus image fusion network. Among them, the main task network and the multi-focus image fusion network of cross-modal image fusion use the same basic skeleton network to share their respective weights, which is used to learn the common features of multiple data. After the basic backbone network, they will follow their own branch networks to learn the ontology characteristics of different data. In branch networks, regularization terms are formed by the characteristics of different tasks.  \textit{On the one hand, over fitting problems caused by experience loss can be prevented; on the other hand, convergence speed can be accelerated by the constraints of different task loss functions.} The method of image fusion proposed by us needs to be completed in the following four steps. \textbf{Firstly}, we propose a theoretical method of constructing nonlinear fusion and feature selection. \textbf{Secondly}, we analyze the loss function of image fusion commonly used. \textbf{Thirdly}, we explore the learning mechanism of human brain for new knowledge, and study the influence of auxiliary task learning mechanism on the main task of image fusion. \textbf{Finally}, we build an semi-supervised multi-task auxiliary learning image fusion framework based on the image fusion method of multi-task loss auxiliary learning optimization.

\subsection{Nonlinear and feature selection modeling}
In this subsection, we will analyze the nonlinear and feature selection characteristics of human visual system by mathematical modeling.

\subsubsection{Feature selection characteristic}
We suppose that the long and wide channels obtained by residual convolution after previous fusion are $Width$ $\times$ $Height$ $\times$ $Channel$ feature graphs $F= [f_1, f_2, ...,f_k,... f_n]$. As shown in formula \ref{gs4} , the global average pooling (GP) operation is performed on the $T_k$ feature map to obtain the global receptive field corresponding to the feature map, so that the network can exclude the spatial relationship between different channels and focus on learning the nonlinear relationship between different feature channels. After GP operation, the convolution, relu activation function, convolution, sigmoid activation function and dot product operation are used to get the output of the attention module. The mathematical model is defined as:

\setlength\abovedisplayskip{1pt}

\begin{equation}
C_{AM}=S(w_2,R(w_1,F_{GP}))*T_k,
\label{gs4}
\end{equation}
Where $C_{AM}$ indicate channel attention module; $S$ and $R$ represent the activation functions of sigmoid and relu respectively; $w_1$ and $w_2$ represent the weight of two convolutions respectively; $F_{GP}$ indicates the output of the input image after GP operation.

\subsubsection{Nonlinear fusion characteristic analysis}
In the task of image fusion, the current commonly used fusion criteria are weighted average, maximum and sum \cite{Ma2018Infrared}, while the research on nonlinear fusion theory is quite limited. But as a highly complex nonlinear system, human brain needs to deal with very complex logical relations when facing various tasks, which can not be expressed by simple weighted average, maximum or principal component. At the same time, fixed image fusion criteria will seriously reduce the generality of image fusion algorithm. Based on this problem and combined with the strong nonlinear fitting ability of the deep convolution neural network, we construct the deep convolution neural network with the characteristics of feature selection to fit the nonlinear weight of image fusion. Our nonlinear fitting network is defined as:
\begin{equation}
f_{Nolinear} =\sum_{i=1}^{n}{(W_i*I_i+W_{i+1}*I_{i+1})},
\label{max}
\end{equation}
Where $W_i$ is our nonlinear fusion weight map; $I_i$ and $I_{i+1}$ represent the image to be fused. We can find that both maximum fusion, weighted average fusion and summation fusion can be regarded as a special case of nonlinear fusion weight. Taking the fusion of two images as an example, the maximum fusion can be regarded as the problem that the weight value is 1 or 0, while the sum can be regarded as the problem that both fusion weights are 1, and the weighted average can be regarded as the fusion weight is 0.5. The reason why the existing algorithm adopts maximum value, weighted average or sum is that it introduces some prior knowledge to some extent. This fusion criterion of artificial design is only robust to specific tasks, and to a certain extent limits the self-learning ability of the network model. \textit{The nonlinear fusion method proposed by us is just to let the network automatically learn fusion weights according to the commonness of image training data and the characteristics of each image task}. The existing image fusion methods lack the exploration of network self-learning, and more specifically specify fusion criteria to improve the accuracy of a specific task. Therefore, we use the mechanism of the human brain auxiliary learning combined with nonlinear convolutional neural network for further research and exploration. \textit{Our proposed image fusion method can learn not only the common characteristics of different data distribution, but also the characteristics of specific data sets}.

\subsection{Loss analysis of image fusion}
\begin{figure*}[ht]
	\centering
	\includegraphics[width=0.8\textwidth]{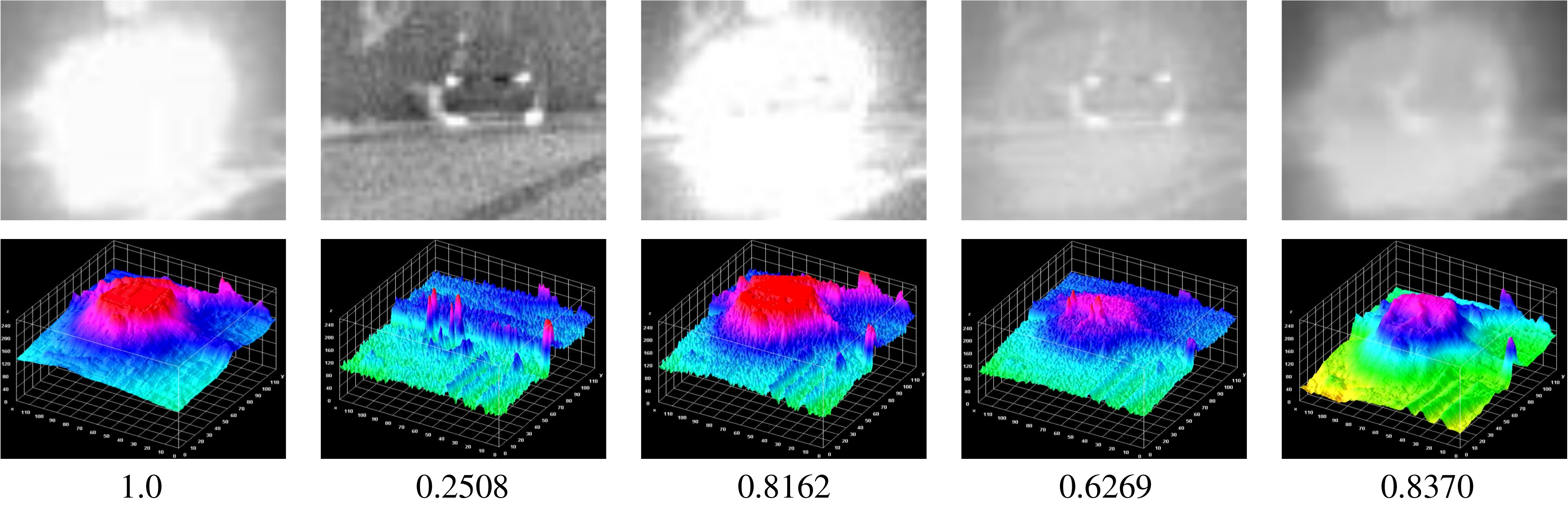}
	\caption{Image quality and structure similarity index.\label{light}}
\end{figure*}

The first image fusion methods described in \ref{method} mainly uses CNN's strong feature representation ability and ignores CNN's strong nonlinear relationship fitting ability. The image fusion methods of end-to-end convolution neural network makes better use of the feature representation ability and relationship fitting ability of convolution neural network, so it will have better performance in generality and robustness than the first method of \ref{method}. The end-to-end deep convolution neural network can mainly be divided into supervised learning and unsupervised learning. In the task of image fusion, the end-to-end image fusion network based on unsupervised learning is mainly the unsupervised learning method proposed by Prabhakar and Yan \cite{PrabhakarK.Ram2017DADU,YanXiang2018UDMI}. This method only uses the combination of single structural similarity index measure (SSIM) \cite{1284395} and variance as the loss function of unsupervised network, and constructs an end-to-end unsupervised learning method. However, a single image quality evaluation parameter can not effectively evaluate image quality. As shown in Fig. \ref{light}, the image comes from FLIR data set \cite{FLIR}. For qualitative analysis of severely degraded image, the SSIM \cite{1284395} value of the image with high subjective score is very low due to the serious degradation of the original image.

From the perspective of cognitive psychology \cite{Treisman1980A}, \textit{this is mainly affected by the visual masking characteristics and brightness contrast characteristics of human visual perception system.}  When the image quality is seriously degraded, SSIM and human subjective evaluation will have a big gap \cite{Treisman1980A}. At the same time, in the image fusion task and image reconstruction task, mean-square-error (MSE) is also a mainstream image fusion objective function. This method can not effectively capture the perception difference between the predicted image and the real image, resulting in the lack of high-frequency information of the reconstructed image, and the image is too smooth \cite{LedigChristian2017PSIS}. To solve this problem, Johson et al. \cite{JohsonJustin2016PLfR} proposed a method of perception loss, which makes full use of high-level global information and low-level detail information, and effectively overcomes the problem of MSE image blur. Although SSIM \cite{1284395} is more suitable for human visual characteristics than MSE or peak signal to noise ratio (PSNR) because of considering the brightness, contrast and structure information of the image, SSIM still does not perform well in the face of high light or serious blur image degradation. Therefore, in our network architecture, we combine perceptual loss \cite{FerzliR2009ANOI}, MSE loss, SSIM loss and PSNR loss to complete network optimization. 

\subsection{Auxiliary Learning mechanism}

In the task of image fusion, the biggest difference between unsupervised learning network and supervised learning network is the lack of ground truth, so it generally lags behind supervised learning network in training accuracy and training difficulty. \textit{Especially in the task of CVS image fusion or cross-modal image fusion such as infrared and visible images, we are faced with not only the lack of real label data, but also the problem that we have not found a complete and effective evaluation of image quality.}
To solve this problem, there are some related researches in the field of computer vision, such as the image fusion of confrontation generation network \cite{MaFusionGAN}, image quality assessment of deep learning IQA \cite{YanBo2019NDNI}, etc. But these methods also have the same problems when training the network, especially for the task of cross-modal image data fusion, no new theoretical breakthrough has been found. Compared with the single loss function training method, our multi-loss function joint training method has a great improvement in accuracy, but the quality of image fusion across data sets is still not better than the supervised learning method and some traditional fusion operators. By analyzing that the process of human brain learning perception for new tasks is based on the perceived task knowledge to assist learning the characteristics of new tasks. Therefore, based on the unsupervised learning network framework proposed in this section, we introduce image reconstruction task and multi-focus image fusion task to assist the learning of cross-modal image fusion task. Through different auxiliary task learning, we can fully mine the hidden features that unsupervised fusion network learning can not get \cite{LiuShikun2018EMLw}. Therefore, we take the image reconstruction task and the multi-focus image fusion task as the object. Based on the unsupervised learning cross-modal image fusion framework, we expand the research on the loss of auxiliary task and propose a auxiliary learning attention network ALAN.

\begin{figure*}[ht]
	\centering
	\includegraphics[width=0.7\textwidth]{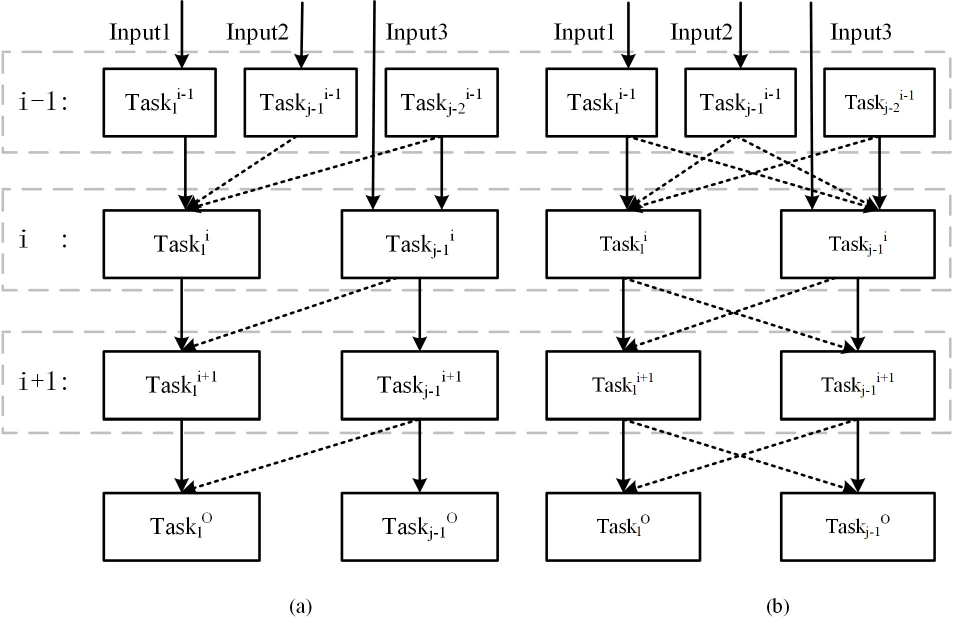}
	\caption{Principles of auxiliary learning process \cite{RusuAndrei2016PNN}\label{Fig5}. Subfigure (a) shows the single task auxiliary primary task auxiliary learning process. Subfigure (b) shows the process of multi-task collaborative optimization.}
\end{figure*}

As shown in Fig. \ref{Fig5}, it is the schematic diagram of our subtask collaborative work. Where (a) shows the process of subtask assisting main task; (b) it is a collaborative optimization process of subtask and main task at the same time. Through such a network structure design, our main network can effectively retain the unique characteristics of two subtasks while extracting its own data characteristics, so as to improve the universality and robustness of the network model.
In order to express the structure of the network more clearly, we build a mathematical model, which is defined as:
\begin{equation}
\label{gs10}
Task_{i}^{(l)}=R\left(W_{i-1}^{(l)} x_{i-1}^{(l)}+\sum_{j<l} W_{i-1}^{(j)} x_{i-1}^{(j)}\right),
\end{equation}
where $Task_{i}^{(l)}$ represents task $0$-$l$ at layer i; R represents the nonlinear activation layer; $W_{i-1}^{(l)}$ represents the convolution weight of task $l$ in layer $i-1$ network; $x_{i-1}^{(l)}$ represents the input of task $l$ in the $i-1$ network; $\sum_{j<l} W_{i-1}^{(j)} x_{i-1}^{(j)})$ represents the sum of $l-1$ tasks in layer $i-1$ convolution neural network.

The task of image reconstruction and multi-focus image fusion is closely related to the task of cross-modal image fusion. By introducing the task of image reconstruction and multi-focus image fusion for auxiliary learning, the end-to-end unsupervised learning network framework can be transformed into a supervised learning network. In the image reconstruction task and the multi-focus image fusion task, because there are many related open data sets with labels, we use the end-to-end supervised learning method to build the deep convolution neural network. In the subtask of image fusion auxiliary learning, we comprehensively consider the structure information, contrast information, brightness information and deep feature information of different network depths. Our loss function is mainly composed of perceptural loss, SSIM loss, MSE loss and PSNR loss. Our loss function is defined as:

\begin{equation}
\label{gs11}
L=\alpha_1 \times L_{SSIM}+\alpha_2 \times L_{PSNR}+ \alpha_3 \times L_{P}+ \alpha_4 \times L_{MSE},
\end{equation}
Where $\alpha_1, \alpha_2, \alpha_3, \alpha_4$ represent the weight of perceptural loss, SSIM loss, MSE loss and PSNR loss respectively.

\subsection{Semi-supervised multi-task network}
Our proposed Semi-supervised learning network framework is shown in Fig. \ref{general}. The network framework mainly includes one main task module and two subtask modules. The main task of image fusion is multi-modal image fusion network. Subtask 1 is the image reconstruction network, which mainly completes the learning process of image hidden features, and uses the hidden features in the main task of cross-modal image fusion. Subtask 2 is the multi-focus image fusion task, which transfers the hidden features learned from the multi-focus data set to the main task of cross-modal image fusion, and promotes the optimization of the main task of image fusion. Combined with two subtask models, the network structure of multi-task auxiliary learning is formed.

\begin{table}[!h]\centering\small
	\renewcommand \arraystretch{1.1}
	\caption{Subtask 1 network parameters.}
	\label{table1}
	\begin{tabular}[b]{p{0.5cm}p{0.8cm}p{1.5cm}p{0.8cm}p{0.8cm}p{1.3cm}}
		\hline
		Type & Kernel & Input &  Out & Stride &Activation \\ 
		\hline
		C1   	 &3&1&16&1&Relu\\
		EC1	 &3&16&16&1&Relu\\
		EC2	 &3&32&16&1&Relu\\
		EC3	 &3&48&16&1&Relu\\
		DC1	 &3&64&64&1&Relu\\
		DC2	 &3&64&32&1&Relu\\
		DC3	 &3&32&16&1&Relu\\
		DC4	 &3&16&1&1&Relu\\
		\hline
	\end{tabular}
\end{table}

\begin{table}[!h]\centering\small
	\renewcommand \arraystretch{1.1}
	\caption{Main task and subtask 2 network parameters.} 
	\label{table3}
	\begin{tabular}[b]{p{0.5cm}p{0.8cm}p{1.5cm}p{0.8cm}p{0.8cm}p{1.3cm}}
		\hline
		Type & Kernel Size&Input & Output & Stride &Activation \\ 
		\hline
		C2   &3&1&64&1&Relu\\
		C3	 &3&1&64&1&Relu\\
		C4	 &3&64&64&1&Relu\\
		C5	 &3&64&64&1&Relu\\
		C6	 &3&256,128,64&64&1&Relu\\
		C7	 &3&256,128,64&64&1&Relu\\
		C8	 &3&64&64&1&Relu\\
		C9	 &3&64&64&1&Relu\\
		C10	 &3&256&1&1&Relu\\
		C11	 &3&256&1&1&Relu\\	
		C12	 &1&64&64&1&Relu\\
		C13	 &3&64&64&1&Relu\\
		C14	 &5&64&64&1&Relu\\	
		C15	 &1&192&64&1&Relu\\
		C16	 &3&192&64&1&Relu\\
		C17	 &5&192&64&1&Relu\\	
		C18	 &1&192&64&1&Relu\\		
		\hline
	\end{tabular}
\end{table}

\subsubsection{Cross-modal image fusion main task}
The main task is composed of multi-scale convolution block (MSRB) \cite{LiJ.2018Mrnf} and channel attention module (CAM) \cite{hu2017squeezeandexcitation}. MSRB can effectively mine hidden features in different scales. Through the CAM and nonlinear convolution neural network, we can effectively simulate the feature selection characteristics and nonlinear combination characteristics of human brain image fusion.

In the main task training process, we used 4000 original CVS data sets, with a single resolution of 1280x1024, and expanded to 20000 through data enhancement. In order to increase the diversity of data, we added 2999 preregistered infrared and visible images in the data set, with a single resolution of 320x256, and expanded to 20000 after data enhancement. In the main task training, all our image inputs are gray-scale image, and the image size is 80x64. Our network training learning rate is 0.0001, batchsize is 16, epoach is 10.

\subsubsection{Image reconstruction subtask}
Subtask 1 is image reconstruction task \cite{Li2018DenseFuse}, which uses end-to-end supervised learning network. Some of the existing image reconstruction tasks, such as super-resolution reconstruction \cite{LedigChristian2017PSIS}, image restoration task \cite{QianRui2017AGAN}, only take down samples of the image, or only consider a specific noise to enhance the restoration operation, so these methods have good performance in a specific data set, but the adaptability is poor in a complex real environment.

In this subtask, our network structure adopts the densenet network framework. For detailed parameters, please refer to \cite{Li2018DenseFuse} and \ref{table1}. On the basis of COCO2014 \cite{LinT.-Y.2014MCCo} data set, we make joint random adjustment of brightness, ambiguity and Gaussian noise in a certain range, so that the distribution of training data is as close as possible to that of real environment data. In the image reconstruction training stage, we used more than 70000 training sets and 10000 verification sets. Due to the limitation of video memory, we adjust the image size after preprocessing to 256x256. Our network training learning rate is 0.0001, batchsize is 8, epoach is 4.

\begin{figure*}[ht]
	\centering
	\includegraphics[width=0.7\textwidth]{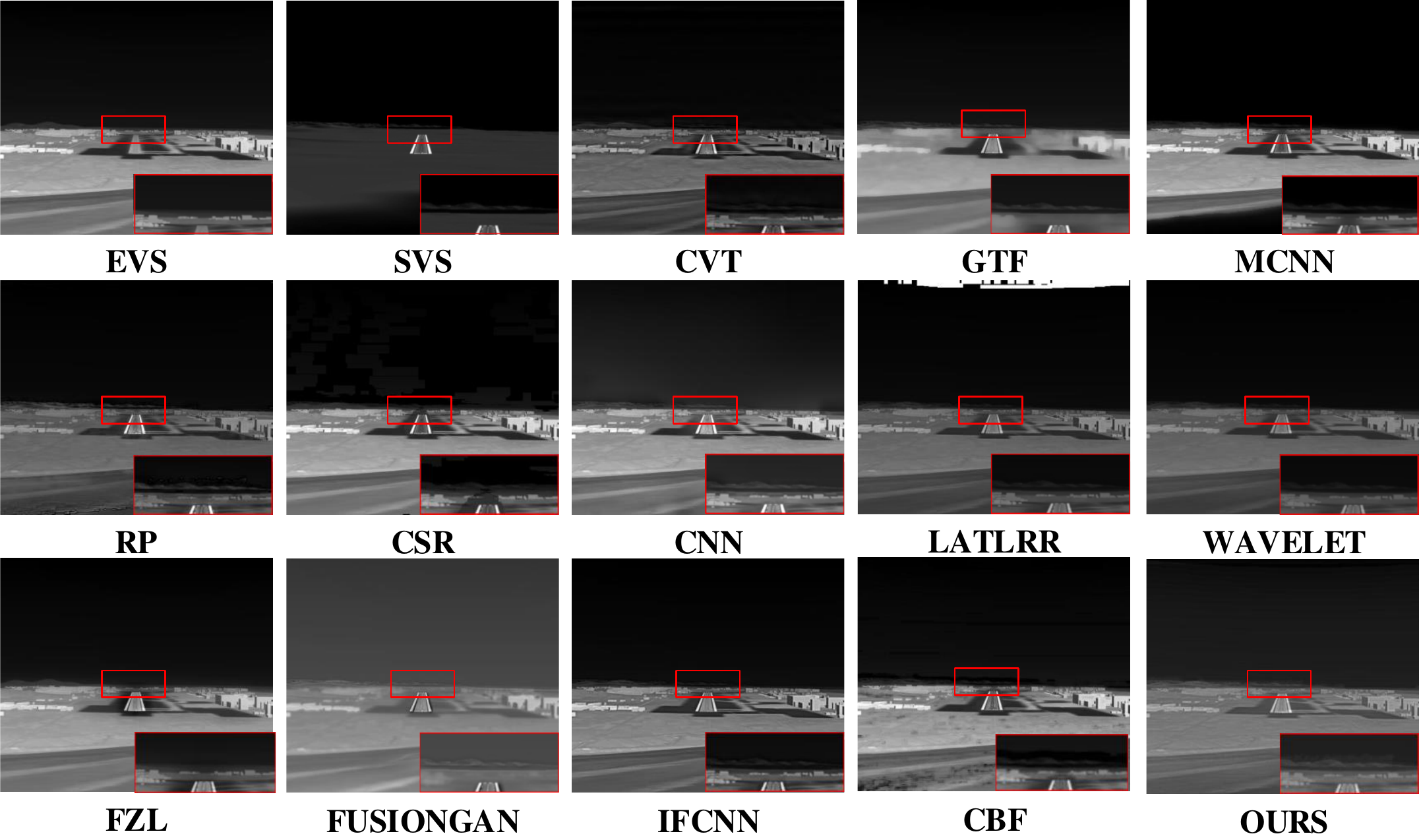}
	\caption{EVS and SVS images with the fusion results obtained by different methods\cite{RusuAndrei2016PNN}. All defined in subsection \ref{setup}\label{f6}.}
\end{figure*}

\subsubsection{Multi-focus image fusion subtask}
Subtask 2 adopts the task of multi-focus image fusion, which adopts the end-to-end supervised learning network framework, which is basically the same as the main task framework. In the training process of the network, we used the data from lytro dataset\cite{Nejati2015Multi}, and we expanded the data to 20000 pieces, with a single resolution of 80x64. Our network training learning rate is 0.0001, batchsize is 16, epoach is 10.

\subsubsection{Training}
In order to avoid the influence of the main task loss function on the convolution weight of subtasks, we train subtasks separately and fix the convolution weight of subtask modules. The convolution weight of subtask and the main task are combined as a part of the basic node of the main task, and the objective function of the main task is used to optimize the weight of the main task node. The whole loss function of the auxiliary learning network is defined as: 

\begin{equation}
L=L_{m}+L_{f}+L_{e},
\label{gs12}
\end{equation}
Where $L$ indicates our network loss function; $L_{m}$ represents multi-focus image fusion task loss function; $L_{f}$ represents the cross-modal image fusion task loss function; $L_e$ indicates image reconstruction task loss function.

In order to display the network detail parameters as shown in Fig. \ref{general} more accurately, we draw Tables 1 \cite{Li2018DenseFuse} and 2. At the same time, in Table 2, the inputs of C6 and C7 are also different due to different fusion criteria F. If the concat fusion criterion is used, the input is 256. If the hybrid fusion criterion is adopted, such as weighted average first and then concat fusion operation, the input is 128. If the additive sum or weighted average fusion criterion is used, the input is 64. If this part of content is modified, the main task network needs to be retrained.

%
%
%
%
\section{EXPERIMENTS}
In this section, experimental setup are presented and comparative experiments result produced along with relevant explanations and analysis experiment are presented.
\label{test}
\subsection{Experiments Setup}
In this section, datasets, metrics and methods for experimental evaluation are presented. At last, implementation details of evaluated methods are introduced.
\subsubsection{Datasets}

\textit{1) TNO \cite{Toet2014}}: It contains multi-spectral (enhanced vision, near infrared and long wave infrared or thermal) night images of different military related scenes, registered in different multi-band camnera systems. There are 21 pairs of image pairs commonly used in existing image fusion algorithms.

\textit{2) Multi-focus \cite{ZhangYu2017Bfbm}}: There are 20 pairs of image pairs commonly used in existing image fusion algorithms. The data involves various scenes, including indoor, outdoor, plant, animal, etc., all images have been registered.

\textit{3) CVS (OURS)}: The dataset is specially used in the field of aviation visual navigation, including synthetic visual image and enhanced visual image. The dataset includes 4000 pairs of original images. CVS image obtained by fusion of EVs and SVS. As there is no public dataset in this field, we rely on the Aeronautical Research Institute to obtain this dataset. Later, we will make and publish the benchmark of the dataset after obtaining the consent of the agency.


\subsubsection{Metrics}
In order to qualitatively evaluate the performance of different algorithms, we mainly use five commonly used objective evaluation indexes of image quality. Cumulative probability of blur detection (CPBD) \cite{NarvekarN.D2011ANIB}, just perceptible blur based on human vision (JNB) \cite{FerzliR2009ANOI}, visual information fidelity (VIF) \cite{Han2013A}, average gradient  (AG) \cite{Cui2015Detail}, SSIM \cite{1284395}. For the specific definition of the above indicators, please refer to the corresponding articles.

\subsubsection{Methods}
We will compare experiments with 16 mainstream algorithms such as fast zero learning (FZL) \cite{Lahoud2019FastZERO}, convolutional sparse representation (CSR) \cite{Liu2016ImageCSR}, deep learning (DL) \cite{Li_2018DL}, dense fuse (DENSE) \cite{Li2018DenseFuse}, generative adversarial network for image fusion (FUSIONGAN) \cite{Ma2018Infrared}, laplacian pyramid (LP)
\cite{Burt1987TheLP}, dual-tree complex wavelet transform (DTCWT) \cite{Liu2015MultiDSIFT}, latent low-rank representation
(LATLRR) \cite{Li2018InfraredLTLRR}, multi-scale transform and sparse representation (LP-SR)
\cite{Liu2015ALPSR}, dense sift
(DSIFT) \cite{Liu2015MultiDSIFT}, convolutional neural network (CNN) \cite{Liu2017InfraredCNN}, curvelet transformation (CVT)
\cite{Nencini2007RemoteCVT}, bilateral filter fusion method (CBF) \cite{Shreyamsha2015ImageCBF}, cross joint sparse representation
(JSR) \cite{Zhang2013Dictionary}, joint sparse representation with saliency detection (JSRSD) \cite{Liu2017InfraredJSR-SD}, gradient transfer fusion (GTF) \cite{Ma2016InfraredGTF}, weighted least square optimization (WLS) \cite{Ma2017InfraredWLS}, a ratio of low pass pyramid(RP) \cite{Toet1989ImageRP}, wavelet \cite{Chipman1995Wavelets}, multi-focus image fusion with a deep convolutional neural network (MCNN) \cite{Liu2017MultiCNN}, image fusion convolution neural network(IFCNN) \cite{ZhangYu2020IAgi}, OURS.

\subsubsection{Implementation details}

\label{setup}

\begin{figure*}[ht]
	\centering
	\includegraphics[width=0.7\textwidth]{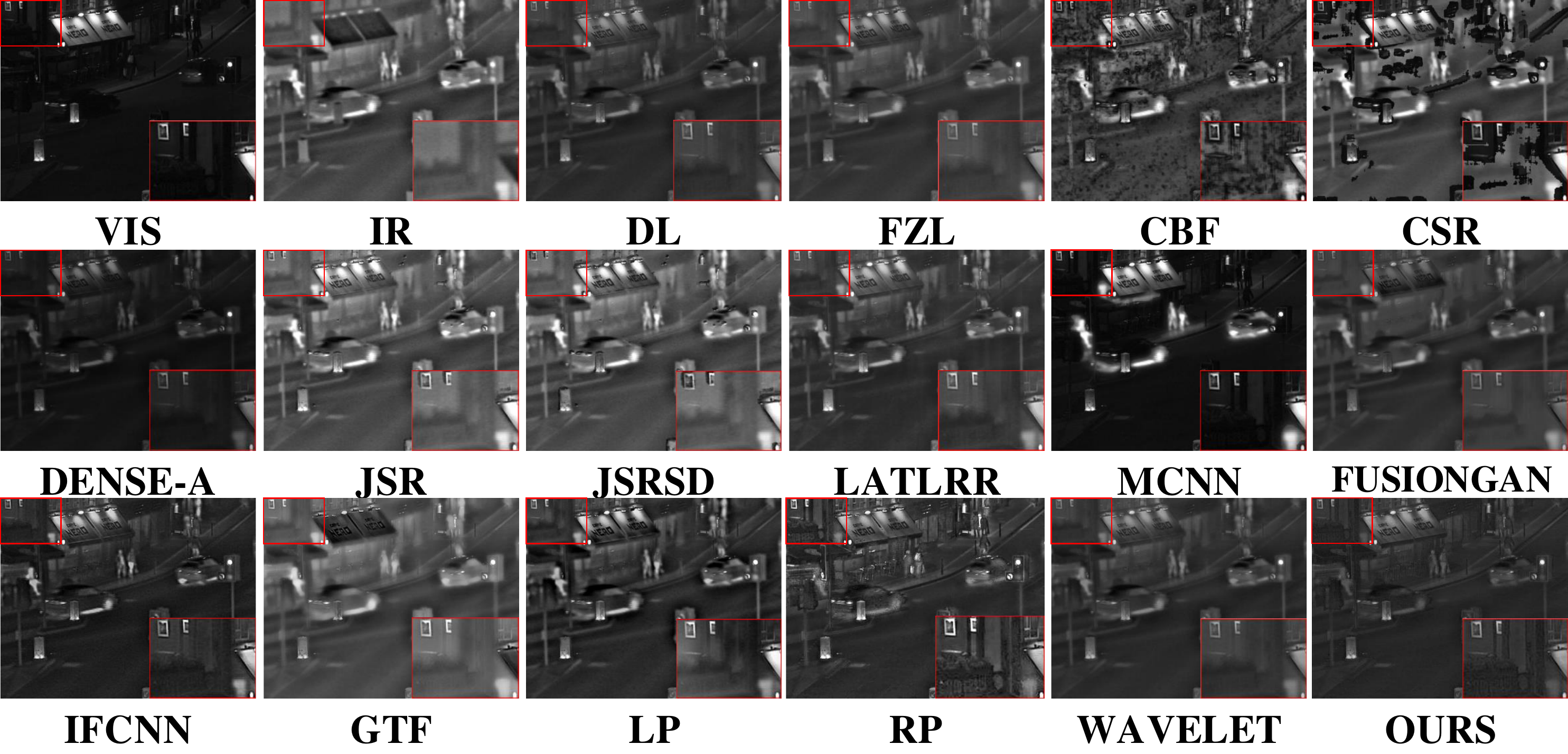}
	\caption{Qualitative fusion results on Visible and thermal infrared images from TNO data set. All defined in \ref{setup}\label{f7}.}
\end{figure*}


\begin{figure*}[ht]
	\centering
	\includegraphics[width=0.7\textwidth]{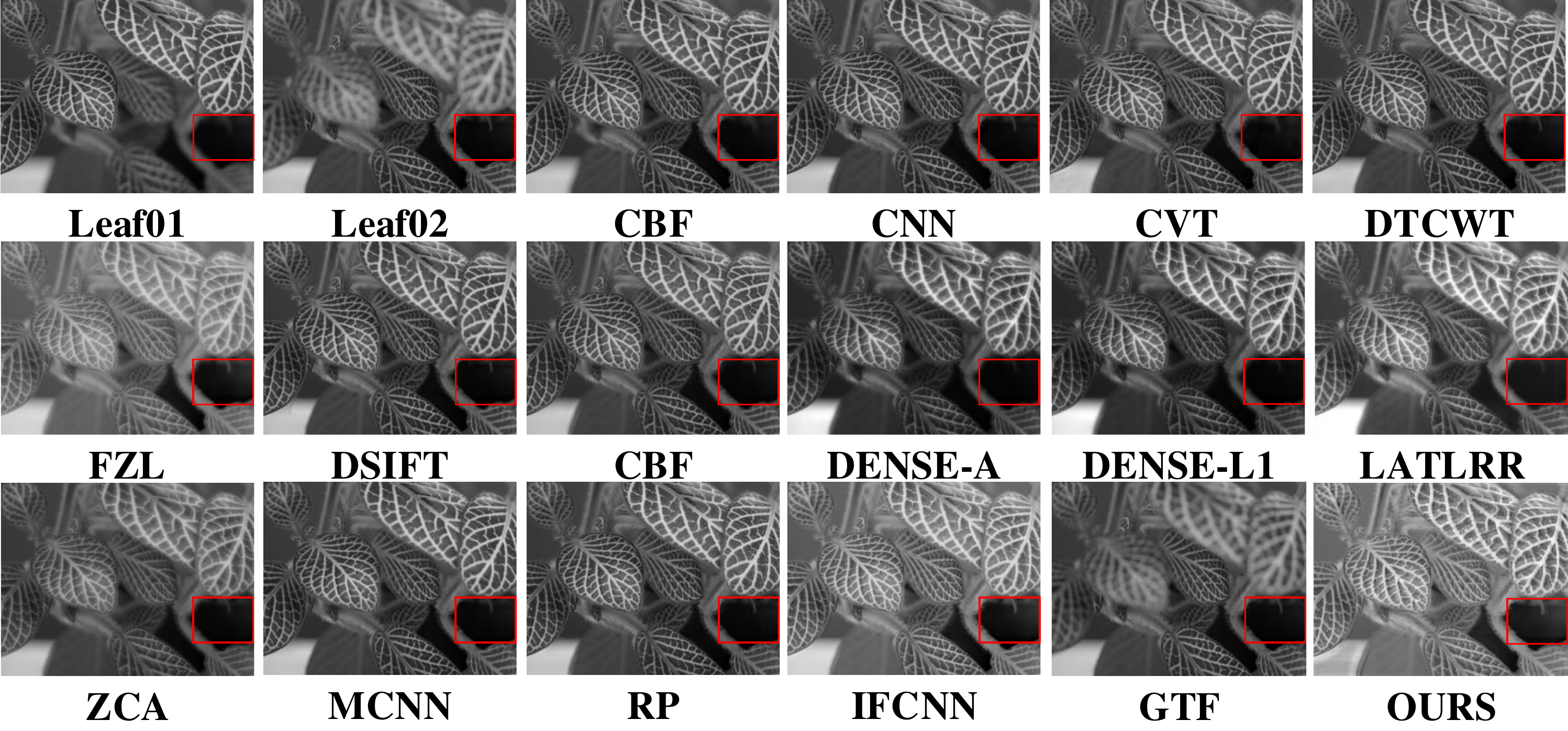}
	\caption{Qualitative fusion results on multi-focus images by different method. All defined in \ref{setup} \label{f8}.}
\end{figure*}

In all experiments, we transform all images into gray-scale images for subsequent image fusion training. At the same time, we need to explain that in all subsequent experiments, our algorithm does not manually adjust parameters for fixed data sets. For different experiments, there will be some changes in the related algorithm experiments, and the changes will be explained in the respective experimental chapters. These algorithms have already published their code, and the relevant algorithm parameters are the same according to the settings in the public paper, and after the publication of our paper, the relevant code and data of our paper will be published on GitHub. Our experimental platform is desktop 3.0 GHZ i5-8500, RTX2070, 32G memory.

\subsection{Comparative experiments}
In this section, in order to verify the robustness of our algorithm, we will evaluate subjective quality and objective quality on TNO data set, CVS data set and multi-focus data set. 

\subsubsection{Combined vision system image fusion experiment}
In CVS image fusion data set, we use some image fusion algorithms shown in \ref{setup} to analyze the CVS images qualitatively. As shown in Fig. \ref{f6}, in the EVS image, due to the influence of dark light, many image texture details existing in the dark light are almost imperceptible to the naked eye, and the existing image fusion algorithms are unable to recover these details well during image fusion. Although RP algorithm and CNN algorithm recover some details, but also introduce some non image information. Compared with other algorithms, our algorithm has a very clear edge detail in the dark part.

\subsubsection{Infrared and visible image fusion experiment}
In this subsection, we will compare experiments on TNO data set.
\begin{table*}[!h]
	\centering\small
	\renewcommand \arraystretch{1.1}
	\caption{Subjective image quality assessment.}
	\label{table4}
	\begin{tabular}[b]{p{1.2cm}p{0.8cm}p{0.8cm}p{0.8cm}p{0.8cm}p{0.8cm}p{0.8cm}p{0.8cm}p{1.6cm}p{0.8cm}p{0.8cm}p{0.8cm}}
		\hline
		DATA & FZL & CSR & CVT & CNN & GTF & LATLR & RP & FUSIONGAN & IFCNN& MCNN & OURS \\ 
		\hline
		CVS  &4.48 & 4.46 & 4.41 & 4.64 & 3.70 & 4.47 & 4.65  & 3.80  & \textbf{4.68}  & 4.40&\textbf{4.76} \\ 
		IR   &4.18& 3.74& 4.51 & 4.19 & 4.08 & 4.43 & \textbf{4.68}  & 4.22  & 4.63  &4.10 & \textbf{4.79} \\ 
		MF   & 4.20& 4.35& 4.57 & 4.66 & 4.25 & 4.34 & \textbf{4.63}  & 4.46  & \textbf{4.70}  &4.67& 4.65 \\ 
		\hline
	\end{tabular}
\end{table*}

\begin{figure*}[ht]
	\centering
	\includegraphics[width=0.8\textwidth]{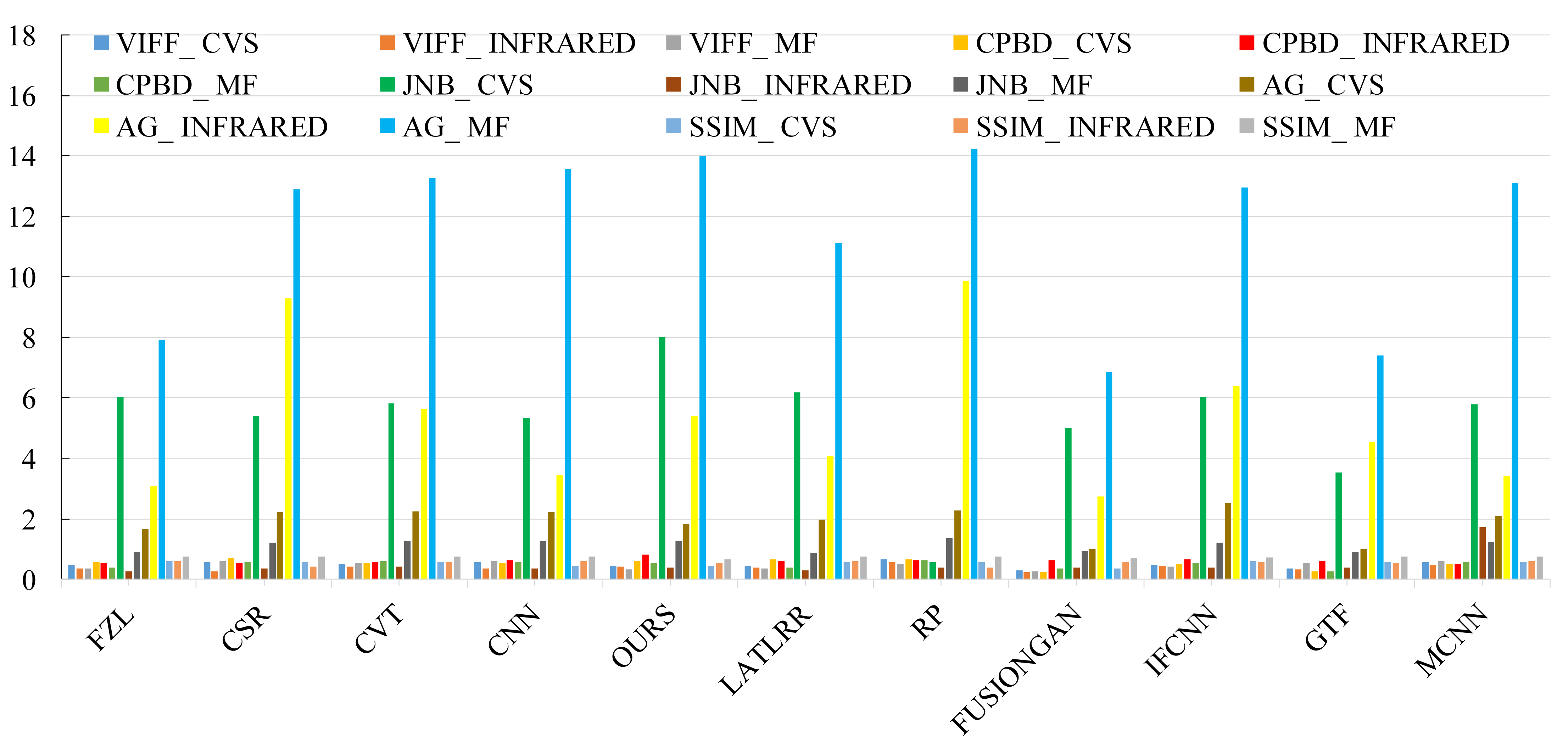}
	\caption{Objective evaluation of image quality on three image fusion datasets \label{f5}. }
\end{figure*}

From Fig.\ref{f7}, we can see that our algorithm can recover more image details while maintaining lower noise compared with other algorithms in infrared and visible image fusion tasks. In this data set, although IFCNN algorithm has higher contrast than our algorithm, but the fusion image of this algorithm also introduces a lot of image noise, which affects the quality of image fusion. The reasons are mainly divided into three parts. Firstly, IFCNN uses supervised learning method to train image fusion, only through data-driven learning to learn the data distribution of multi-focus data sets; then, IFCNN network adds the prior knowledge of human beings, and adopts the maximum fusion criterion for cross-modal infrared and visible images; finally, image quality evaluation is not perfect. Our algorithm adopts semi-supervised learning network framework, which can automatically learn the nonlinear fusion weights of images, rather than the specified fusion rules. 

\subsubsection{Multi-focus image fusion experiment}
At the same time, we have also carried on the correlation experiment verification to many kinds of image fusion algorithms in the multi-focus image data set. Through the analysis of experimental data, we can see that our algorithm has higher entropy value and gradient value than other algorithms in the multi-focus image, which shows that the fused image information is more abundant and the resolution is better. Especially in the case of low illumination, our algorithm can still better recover the texture details of the image, more in line with the human visual perception characteristics.

\subsubsection{Subjective image quality assessment experiment}
From Fig. \ref{f6}, \ref{f7}, \ref{f8}, \ref{f5}, we can find that the image fusion quality of many algorithms is far from the subjective evaluation of human, but the objective evaluation indexes of related images are very high, such as gradient and SSIM. \textit{The main reason is that these algorithms introduce a lot of noise and edge oscillation effects in the process of image fusion}, such as CBF, CSR and IFCNN. \textit{In order to further verify this conclusion, we evaluate the fusion effect image of the above algorithm subjectively, and use mean opinion score (MOS) as the evaluation index of image quality}. MOS uses a 0-5 scoring criterion. The higher the score is, the better the image quality will be. In this experiment, we invited 10 professors, doctors and other professional researchers of computer image processing to participate in the subjective quality evaluation of the fusion image of different algorithms. In the experimental data processing, we remove the MOS highest score and the lowest score, and take the average score of each data set test data as the final evaluation score. The experimental results are shown in Table \ref{table4}. We can see that the ALAN image fusion method proposed by us has better subjective score than other existing algorithms in different image data sets. Compared with other algorithms, IFCNN has a better subjective score in the multi-focus image dataset, mainly because IFCNN uses multi-focus data for supervised learning training. In order to improve the generality of the network, IFCNN directly replaces the fusion criteria with maximum, weighted average and sum, but it is precisely because of the use of supervised learning method that it can not migrate to the image data without labels, which limits the robustness and generality of the algorithm to a certain extent. At the same time, we can also find that compared with the traditional CVT, GTF, RP, OURS, IFCNN, CNN has better accuracy and robustness in multiple datasets.
\begin{figure}[ht]
	\centering
	\includegraphics[width=0.48\textwidth]{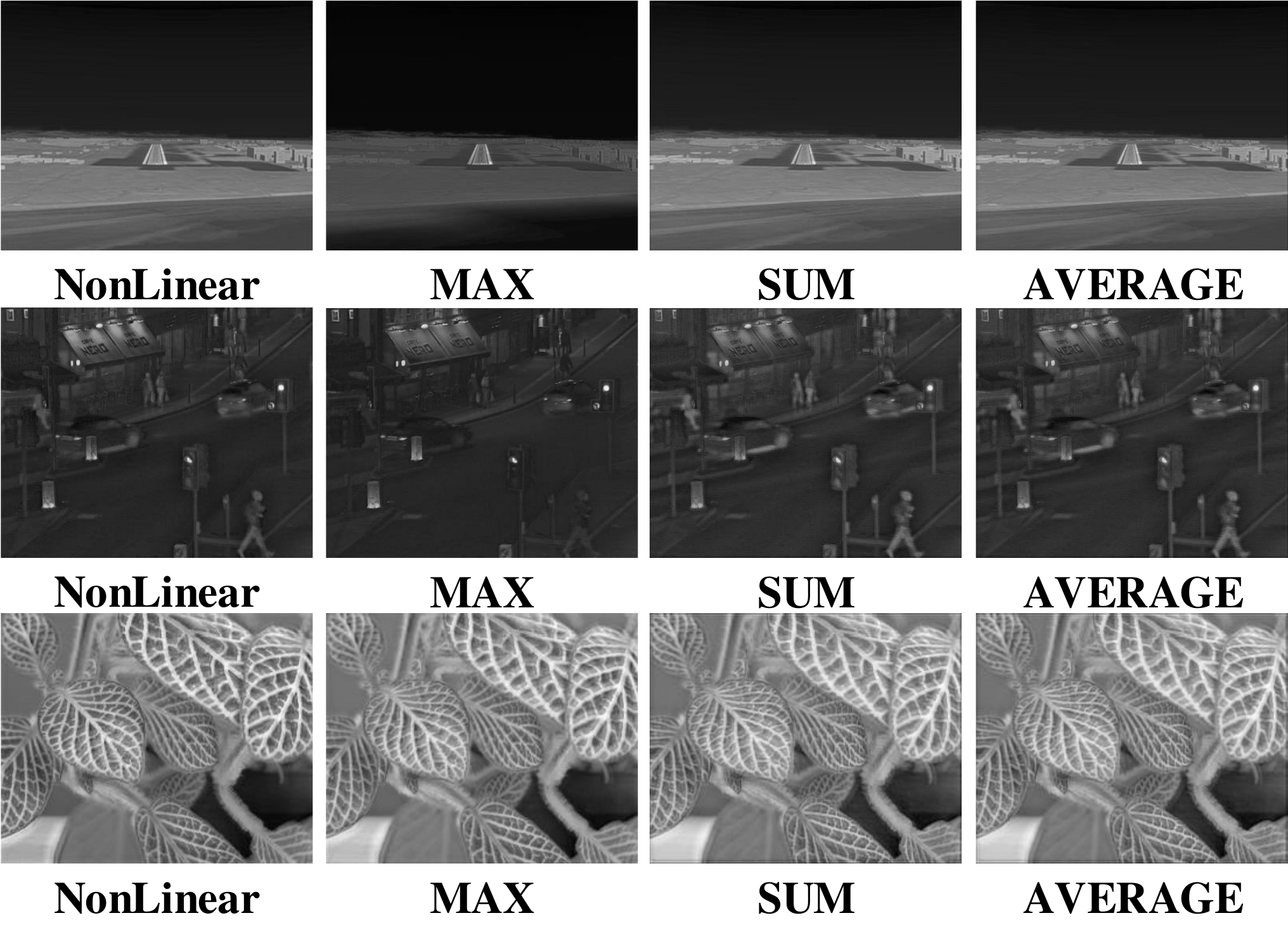}
	\caption{Image fusion effect of different fusion criteria. The first column is the nonlinear fusion criterion, the second column is the maximum fusion criterion, the third column is the sum fusion criterion, and the fourth column is the weighted average fusion criterion \label{f10}.}
\end{figure}

\subsection{Analysis experiments}
At the same time, we also carry out ablation experiments for our method, mainly including comparative analysis experiments for our algorithm using different fusion criteria, and analysis and comparison experiments for single subtask and multi-task loss.

\subsubsection{Comparative experiment of different fusion criteria}
In this experiment, we compare the nonlinear fusion criteria, maximum fusion criteria, sum fusion criteria and weighted average fusion criteria based on our proposed network framework.

From Fig. \ref{f10}, we can see clearly that our nonlinear fusion criteria have very similar fusion effect with sum fusion criteria and weighted average fusion criteria in CVS image fusion data set. The three criteria are well fused in texture details, generally better than the maximum fusion criteria. In infrared and visible image data sets, our method is generally superior to the other three fusion criteria. The sum fusion criterion is very similar to the weighted average fusion criterion, while the maximum fusion captures a large number of features of visible image, and ignores the effective features of visible image, so the performance is poor. In the multi-focus data set, the maximum fusion, sum fusion and weighted average fusion criteria are better in the dark area recovery, but our algorithm has more advantages in overall clarity.

\subsubsection{A comparative experiment of main task fusion algorithm for single fusion task and subtask collaboration}

\begin{figure*}[ht]
	\centering
	\includegraphics[width=0.7\textwidth]{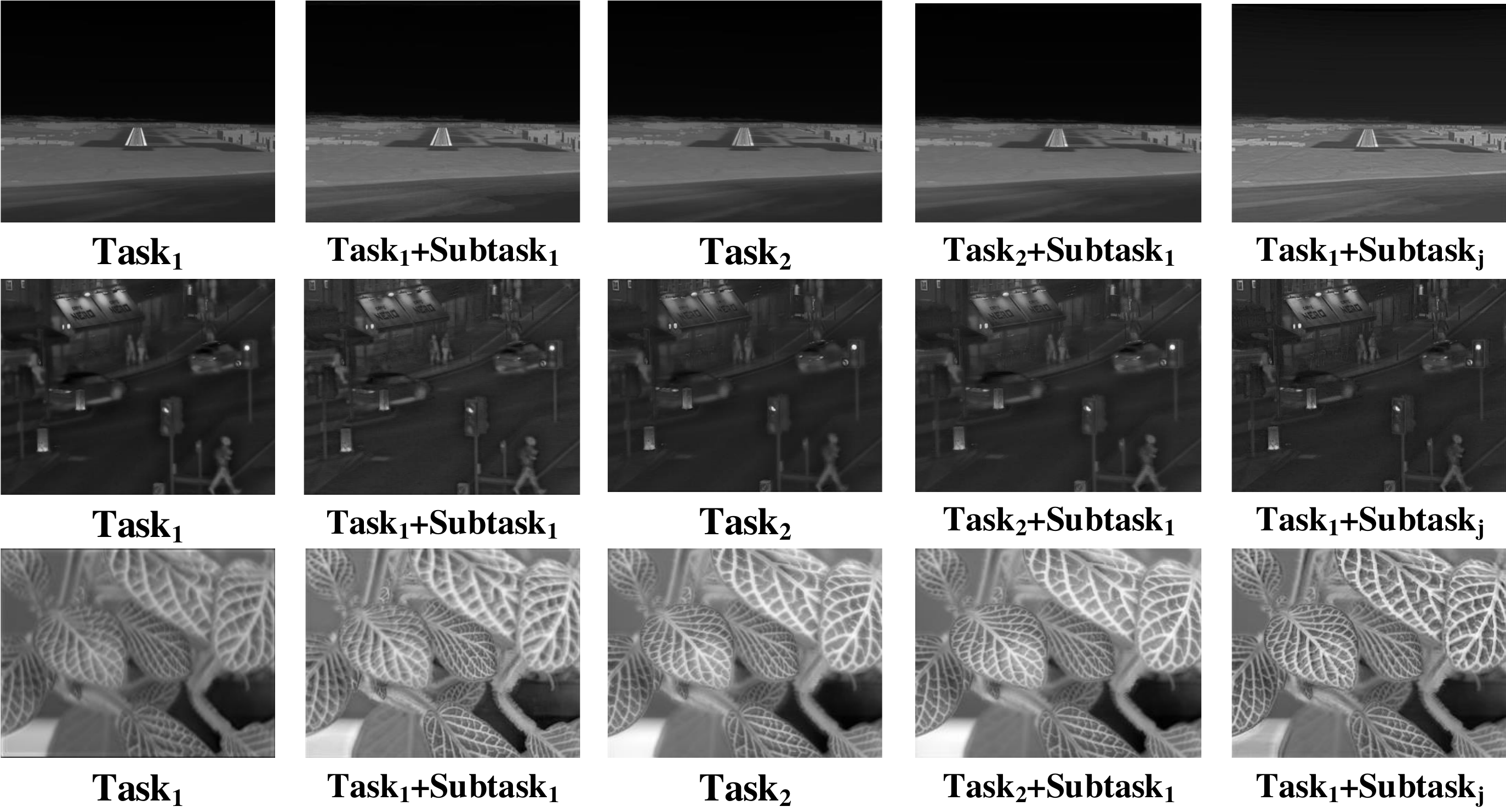}
	\caption{Objective evaluation of image quality. The first column represents a single unsupervised learning image fusion task. The second column shows the fusion effect of unsupervised learning method combined with reinforcement subtask auxiliary learning. The third column represents the rendering of single multi-focus image fusion task. The fourth column shows the image fusion effect of multi-focus image fusion task and image enhancement subtask. The last column is the image fusion effect of multi-task auxiliary learning proposed by us \label{f11}. }
\end{figure*}
From Fig. \ref{f11}, we can clearly see that single image fusion task has better performance in their respective training tasks, but poor performance in cross data. The combination of single image fusion task and image enhancement task can improve the clarity of image to a certain extent. Compared with single unsupervised learning, multi-focus image fusion has a worse performance in the CVS image data set, while in the infrared and visible image data set, it has the opposite performance. In infrared and visible image data set, the effect of single unsupervised learning network combined with image enhancement subtask is much better than that of single unsupervised learning method. Through the auxiliary learning optimization of image enhancement task and multi-focus image fusion task, the universality and robustness of image fusion algorithm on cross dataset can be effectively improved. This is because our image fusion method can effectively extract the common features of multiple data distribution, and also retain some characteristics of different data.

\section{DISCUSSION}
A large number of experiments in section \ref{test} verify that our proposed semi-supervised learning cross-modal image fusion method is better than the existing image fusion method. This also proves the validity of our simulation of human visual characteristics. We think that there are several main reasons. \textbf{Firstly}, the construction of multi-loss function. Compared with the traditional algorithm, deep learning algorithm has a very strong ability of feature representation and feature relationship fitting, but whether the deep network model can learn the subjective intention of human beings and whether the loss function is reasonably constructed through data-driven has an important relationship. However, although there are many methods to evaluate image quality, the existing objective function evaluation methods are relatively single \cite{Ma2018Infrared}. Single image quality evaluation method can not effectively represent image quality, so it is important to study the influence of multi-loss functions on image quality. \textbf{Secondly}, the auxiliary learning mechanism. When single task network training and learning, it is often affected by data noise, insufficient training data, cross-modal and improper loss function, which leads to some hidden features of data can not be learned. Through auxiliary task learning, the learning ability of main task can be effectively optimized. In addition, the network can learn how to learn and reduce the subjective interference through the mechanism of multi-task auxiliary learning. We can let the network automatically learn the common characteristics of different image data and the characteristics of their own image data only through the collaborative optimization of different tasks, which plays an important role in improving the robustness and generality of the network architecture. \textbf{Thirdly}, the nonlinear combination and feature selection of human visual system. Through the effective combination of attention mechanism and nonlinear convolution neural network, we can learn the non mutually exclusive nonlinear fusion weights between cross-modal images. Experiments show that this method is more consistent with the human brain image fusion mechanism. \textbf{Finally}, semi-supervised learning. At present, in the field of image fusion, it is very difficult to obtain the supervised learning labels of both infrared and visible image data set, or CVS image data set. For the main task of image fusion, we do not need ground truth labels. By introducing the auxiliary learning strategy, we can effectively transform the unsupervised learning network into the supervised learning, and effectively improve the robustness and universality of image fusion. At the same time, our image fusion algorithm is only applied to the gray image fusion task, but our algorithm idea can be applied to the color image fusion task. If you want to apply it to the task of color image fusion, you need to refer to \cite{PrabhakarK.Ram2017DADU,ZhangYu2020IAgi,Li2018DenseFuse} to modify the data loading interface and retrain the network.

\section{CONCLUSION}
Based on the three characteristics of human visual perception system, we propose a robust multi-task auxiliary cooperative optimization image fusion method. \textit{The main differences between our image fusion method and existing image fusion method are as follows}. \textbf{Firstly}, our image fusion network adopts multi-loss functions, which can better represent the image quality than the current single loss function method. \textbf{Secondly}, we combine the attention mechanism and the deep convolution neural network to simulate the feature selection and nonlinear combination characteristics of the human visual system. \textbf{Thirdly}, the auxiliary learning mechanism is introduced into the image fusion task, and the main task of image fusion is effectively optimized through multi-tasks. \textit{To a certain extent, it overcomes the difficulty of modeling the loss function of cross-modal image, and provides a new idea for cross-modal image fusion.} \textbf{Finally}, the semi-supervised learning cross-modal image fusion framework proposed by us. It is more robust than the existing algorithms. In addition to the CVS image fusion task, it can also be applied to infrared and visible image fusion task and multi-focus image fusion task. A large number of experiments show that our image fusion method is more robust than the existing mainstream algorithm. Although our algorithm framework does not fully simulate the mechanism of human brain image fusion, our simulation of the characteristics of human brain image fusion mechanism is consistent with the mechanism of human brain image fusion. Our future work is to further study the direction of intelligent image fusion, such as context-awareness, self-learning image fusion methods and so on.


%




\ifCLASSOPTIONcaptionsoff
  \newpage
\fi

\bibliographystyle{IEEEtran}
\bibliography{refs}

\end{document}